%% file: main.tex
\documentclass[sigconf]{acmart}

\usepackage{booktabs} % For formal tables

\usepackage{lipsum}
\usepackage{url}
\usepackage{algorithm}
\usepackage[noend]{algpseudocode}% http://ctan.org/pkg/algorithmicx
\algnewcommand\algorithmicforeach{\textbf{for each}}
\algdef{S}[FOR]{ForEach}[1]{\algorithmicforeach\  #1\ \algorithmicdo}

\usepackage{todonotes}
\usepackage{amsmath}
\usepackage{microtype}
\newcommand{\floor}[1]{\left\lfloor #1 \right\rfloor}

\usepackage{enumitem}
\DeclareMathOperator*{\argmax}{arg\,max}

\newcommand\algoname{ME-MAP-Elites}

\copyrightyear{2021} 
\acmYear{2021} 
\setcopyright{acmlicensed}\acmConference[GECCO '21]{2021 Genetic and Evolutionary Computation Conference}{July 10--14, 2021}{Lille, France}
\acmBooktitle{2021 Genetic and Evolutionary Computation Conference (GECCO '21), July 10--14, 2021, Lille, France}
\acmPrice{15.00}
\acmDOI{10.1145/3449639.3459326}
\acmISBN{978-1-4503-8350-9/21/07}

\begin{document}
\title{Multi-Emitter MAP-Elites}
\subtitle{Improving quality, diversity and data efficiency with heterogeneous sets of emitters}

\author{Antoine Cully}
\orcid{0000-0002-3190-7073}
\affiliation{%
  \institution{Adaptive and Intelligent Robotics Lab,\\Department of Computing, Imperial College London}
  \city{London} 
  \postcode{SW7 2AZ}
  \country{United Kingdom}
}
\email{a.cully@imperial.ac.uk}

\renewcommand{\shortauthors}{A. Cully}

\begin{abstract}
Quality-Diversity (QD) optimisation is a new family of learning algorithms that aims at generating collections of diverse and high-performing solutions. Among those algorithms, the recently introduced Covariance Matrix Adaptation MAP-Elites (CMA-ME) algorithm proposes the concept of emitters, which uses a predefined heuristic to drive the algorithm’s exploration. This algorithm was shown to outperform MAP-Elites, a popular QD algorithm that has demonstrated promising results in numerous applications. 
In this paper, we introduce Multi-Emitter MAP-Elites (ME-MAP-Elites), an algorithm that directly extends CMA-ME and improves its quality, diversity and data efficiency. It leverages the diversity of a heterogeneous set of emitters, in which each emitter type improves the optimisation process in different ways. A bandit algorithm dynamically finds the best selection of emitters depending on the current situation.
We evaluate the performance of ME-MAP-Elites on six tasks, ranging from standard optimisation problems (in 100 dimensions) to complex locomotion tasks in robotics. Our comparisons against CMA-ME and MAP-Elites show that ME-MAP-Elites is faster at providing collections of solutions that are significantly more diverse and higher performing. Moreover, in cases where no fruitful synergy can be found between the different emitters, ME-MAP-Elites is equivalent to the best of the compared algorithms. 

\end{abstract}




\maketitle

\input{body}

%\newpage

\bibliographystyle{apalike}
\bibliography{biblio} % replace by the name of your .bib file

\end{document}

%% file: body.tex
\section{Introduction}
\begin{figure}[!t]
\centering \includegraphics[width=\columnwidth]{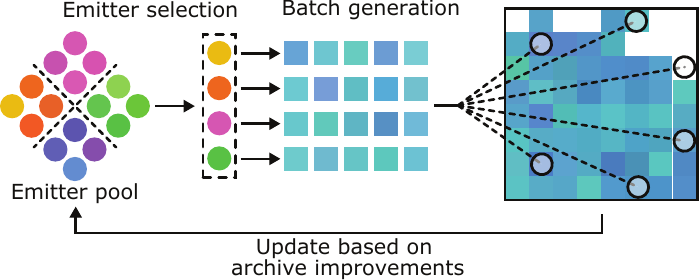}
\caption{High-level concepts of \algoname{}. A set of active emitters is selected from a heterogeneous pool of emitters. The active emitters are used to generate a batch of solutions to be evaluated and potentially added to the archive following the usual MAP-Elites procedure. The outcomes (proportion of successes) of the additions are used as a reward signal for a multi-arm bandit algorithm to bias the emitter selection toward successful emitter types.}
\label{fig:concept}
\end{figure}

Every day, learning algorithms play a more important role in our society, from controlling the efficiency of data centres \citep{evans2016deepmind} to suggesting medical treatments \citep{komorowski2018artificial}. They are also valuable tools to design intelligent and autonomous robots, which have tremendous potential for our society. For instance, robots can substitute for humans in extremely dangerous conditions, such as mining operations, space exploration, or search and rescue missions to find survivors after natural or man-made catastrophes \citep{murphy2014disaster}. Learning algorithms can help robots to discover their capabilities \citep{cully2019Autonomous, paolo2019unsupervised}, or to face unexpected situations like mechanical damage \citep{cully2015robots, bongard2006resilient}. This is crucial, as ``failure of a robot can obstruct the mission execution or cause it to completely fail'' \citep{siciliano2016springer}, which can have dramatic consequences when people's lives is at stake. 

Quality-Diversity optimisation is a new family of learning algorithms that recently emerged from the evolutionary computation community \citep{cully2018quality, pugh2015confronting}. It originates from the concepts of Novelty Search \citep{lehman2011abandoning}.
Its main specificity is to learn, not just one solution, but a large diversity of high-performing solutions. This takes the form of a collection of diverse solutions. This is particularly useful because it can provide alternative solutions to a given problem. For instance, if a robot becomes damaged and the optimal solution it used so far becomes ineffective, the robot can switch to an alternative way to accomplish its mission \citep{cully2015robots}.
This diversity of solutions can also be used to capture the versatility of robots. In this case, it is important to have a collection of diverse solutions in which each one of them is high-performing \citep{cully2015evolving}. Finally, the divergent search capability of Quality-Diversity provides precious stepping stones which can improve the overall optimisation process \citep{gaier2019quality}.

A well-known algorithm of this family is MAP-Elites \cite{mouret2015illuminating}. Like in most learning algorithms, the time (or amount of tests) needed to learn a solution remains a challenge for many applications. For instance, in robotics, every test is particularly time-consuming and bears the risk of damaging the robot. 
To improve this aspect, the recently introduced Covariance Matrix Adaptation MAP-Elites (CMA-ME) algorithm \citep{fontaine2020covariance} proposes the concept of emitters based on CMA-ES\cite{hansen2016cma}, which uses a predefined heuristic to drive the algorithm’s exploration. CMA-ME has been used with emitters aiming at either finding higher-performing solutions (optimising emitter), exploring different regions of the behavioural space (random direction emitter), or improving the content of the collection (improvement emitter).  The two last ones have shown particularly promising results, outperforming MAP-Elites in different settings~\citep{fontaine2020covariance}.

In this paper, we introduce a new algorithm, named Multi-Emitter MAP-Elites (\algoname{}), which focuses on improving the quality, diversity and data efficiency of QD algorithms. \algoname{} extends the concept of emitters from CMA-ME by jointly considering a heterogeneous set of emitters to leverage the strength of each emitter type, while mitigating their relative weaknesses. We also show we can use emitters that are not based on CMA-ES, for instance to replicate the behaviours of MAP-Elites.

We evaluate \algoname{} on six simulated tasks taken from the literature using search spaces ranging from 36 to 100 (real-valued) dimensions, and we compare its performance against a state-of-the-art variant of MAP-Elites \citep{vassiliades2018discovering} and the three existing variants of CMA-ME \citep{fontaine2020covariance}. The experimental results demonstrate that \algoname{} consistently improves the quality (i.e., better fitness of the individuals contained in the archive) and the diversity (i.e., larger coverage of the behavioural space) of the produced archive, with an improved data efficiency. In the most challenging cases, where no fruitful emitter combination can be found, \algoname{} shows similar performance as the best competitor, which is particularly task-dependent.

\section{Related work}
\subsection{Quality Diversity algorithms}
Quality-Diversity (QD) is a new family of algorithms that originates from the concepts of divergent search, like Novelty Search \citep{lehman2011abandoning, lehman2011evolving}. QD aims at generating a collection of diverse solutions that are all as high-performing as possible and as different as possible from each other. This approach has been applied to many domains, such as in robotics to learn a diverse set of high-performing controllers \citep{cully2019Autonomous, vassiliades2018using}, in video-games to generate a variety of dungeons \citep{alvarez2019empowering} or card decks \citep{fontaine2019}, or in workforce scheduling and routing tasks \citep{urquhart2019quantifying}.

The Novelty Search \citep{lehman2011abandoning} and Novelty Search with Local Competition \citep{lehman2011evolving} algorithms are two seminal works that led to the emergence of the Quality Diversity family. They have been followed a couple of years later by the MAP-Elites algorithm \citep{mouret2015illuminating} and the concept of BR-Evolution~\citep{cully2015evolving}, which propose to consider the Novelty archive to be the result of the algorithm rather than its population.
Quality-diversity algorithms optimise a specific type of function that in addition to returning a fitness value, also returns a ``behavioural descriptor'' that is a feature vector used to identify different types of solutions (in MAP-Elites) or compute the novelty of the solutions (in Novelty Search).

MAP-Elites discretises the space of the possible behavioural descriptors into a grid (also called archive), and its goal is to fill each cell of this grid with the highest performing individuals. The algorithm starts with an initialisation phase in which a fixed number (e.g., 500) of solutions are randomly generated, evaluated and then placed into the grid. After the initialisation, MAP-Elites enters in its main loop for a predefined number of generations (e.g., 20k). Each iteration is composed of four steps: 1) uniform random selection of a batch of solutions from the grid, 2) creation of mutated copies of the selected solutions, 3) evaluation of the new solutions and 4) potential addition of the new solutions into the grid. During the evaluation, both the fitness and the behavioural descriptor of each individual are recorded and are used to decide whether or not the individual should be added to the map. The behavioural descriptor determines the cell corresponding to the evaluated solution. If this cell is empty, then the solution is added. Otherwise, a competition between the already present solution and the new one occurs and the one with the highest fitness is kept in the grid.
Throughout the generations, more cells get filled and better solutions are inserted into the grid.

Several works in the literature have introduced elements to improve MAP-Elites on various aspects. For instance, efforts have been made to enable the use of high-dimensional behavioural descriptors, via CVT-MAP-Elites \citep{vassiliades2018using}, AURORA \citep{cully2019Autonomous}, or TAXONS \citep{paolo2019unsupervised}, to ensure adequate discretisation of the behavioural space \citep{fontaine2019} or its robustness to noisy domains \citep{flageat2020fast, justesen2019map}.

\subsection{Improving data-efficiency in QD algorithms }

In this paper, we are mainly interested in the improvement of the data-efficiency by changing the way solutions are sampled at every generation. Works have proposed promising alternative sampling mechanisms. For instance, instead of using a uniform distribution for the random selection, Cully and Demiris~\cite{cully2018quality} proposed to bias the selection according to a ``curiosity score'', which dynamically captures how likely is a solution to produce offspring that will be added in the archive. This approach led to statistically significant, yet limited improvements. Following the same objective, Vassiliades and Mouret~\cite{vassiliades2018discovering} introduced a new variation operator that biases the exploration of the search space towards the hyper-volume of the elites showing higher performance than the other variation operators usually used in MAP-Elites.

The SAIL algorithm \citep{gaier2018data} proposes to use a surrogate model, built on the solutions previously evaluated, to infer the performance of solutions that could be evaluated in the future. This approach is particularly interesting for highly-expensive black-box functions, like in the automatic design of optimised aerodynamic shapes \citep{gaier2017aerodynamic}, as it allows to significantly reduce the number of evaluations by filtering all solutions predicted to be not competitive. This approach assumes that a meaningful surrogate model can be built with a small dataset, which is usually accomplished with a Gaussian Process, but can become challenging in high-dimensional search space or when the fitness landscape is particularly rugged \citep{rasmussen2003gaussian}. 

More recently, Fontaine et al.~\cite{fontaine2020covariance} proposed the concept of ``emitter'' in MAP-Elites. An emitter is responsible for the generation (or emission) of potential solutions to be evaluated. In the standard MAP-Elites framework, an emitter would emit solutions by uniformly selecting a solution from the grid and returning a mutated copy of this solution. Fontaine et al.~\cite{fontaine2020covariance} introduced the Covariance Matrix Adaptation MAP-Elites (CMA-ME) algorithm using more advanced emitter types based on Covariance Matrix Adaptation Evolution Strategy (CMA-ES, \cite{hansen2016cma,hansen2001completely}). Each emitter type follows a CMA-ES optimisation process to sample solutions that are expected to maximise the intrinsic motivation of the emitter, such as improving the overall quality of the MAP-Elites grid, moving in an arbitrary direction in the behavioural space, or maximising the fitness. When an emitter stops making improvements, it is re-instantiated in another region of the search space defined by a solution already in the grid. CMA-ME runs multiple emitters in parallel to explore different regions of the search space at the same time. However, this set of emitters only contains a single type of emitter.

\section{\algoname{} }

\begin{algorithm}
\small
\caption{\algoname{}~( $G$ generations, $N$ active emitter slots, $P$ emitter  types)}
\label{algo:ME-MAP-Elites}
\begin{algorithmic}
\State $\textrm{MAP-Elites-Grid} \leftarrow \emptyset$%\Comment{\emph{Creation of an empty grid.}}
\State $\textrm{emitter\_pool} \leftarrow \emptyset$
\State $\textrm{active\_emitters} \leftarrow \emptyset$
\State $\textrm{successes} \leftarrow zeros(\textrm{len}(N*P))$ %\Comment{\emph{Successes for each emitter}}
\State $\textrm{selection} \leftarrow zeros(\textrm{len}(N*P))$ %\Comment{\emph{Selections for each emitter}}
\vspace{0.2cm}
\State $\textrm{Initialise(emitter\_pool)} $ \\ \Comment{\emph{Initialise emitter pool with $N$ emitters for each of the $P$ types.}}
\vspace{0.2cm}
\For{gen $  = 1\to G$} %\Comment{\emph{The main loop repeats during $I$ iterations.}}
\State $\textrm{emitter\_pool.append(remove\_terminated(active\_emitters))}$ \\ \Comment{\emph{Returns the emitters that have terminated to the emitter pool}}
\State $ \textrm{nb\_needed\_emitters} \leftarrow N -  \textrm{len(active\_emitters)} $
\State $\textrm{sorted\_pool} \leftarrow  \textrm{descending\_sort(emitter\_pool)}$ \\ \Comment{sort based on eq.\ref{eq:ucb}} \Comment{$\textrm{with} \frac{successes[e]}{selection[e]} +\zeta \sqrt{ \frac{\log\sum selection}{(selection[e])}}$}
\State active\_emitters  $\leftarrow$ sorted\_pool.pop([1:nb\_needed\_emitters]) %\\ \Comment{\emph{Select $N$ emitters from the pool}}

\ForEach {emitter $e \in$ active\_emitters}
\State batch $\leftarrow e$.generate\_samples() %\\ \Comment{\emph{use emitter to generate solution batch}}
\ForEach {solution $ x_i \in $ batch}
\State $bd, fit  \leftarrow$ evaluate$(x_i)$
\State added $\leftarrow \textrm{MAP-Elites-Grid.add\_attempt}(x_i,db,fit)$
\State $\textrm{selection}[e] \leftarrow \textrm{selection}[e] + 1$
\If{added}
\State $\textrm{successes}[e] \leftarrow \textrm{successes}[e] + 1$
%\Comment{\emph{Successes for each emitter}}
\EndIf

\EndFor
\EndFor
\EndFor
\State return(MAP-Elites-Grid)
\end{algorithmic}
\end{algorithm}

Multi-Emitter MAP-Elites (\algoname{}) is a direct extension of CMA-ME, in which instead of using exclusively one type of emitter, multiple emitter types are used together in a heterogeneous emitter set. Moreover, the proportion of each emitter type used at each generation is dynamically adjusted by using a bandit algorithm \citep{auer2002finite} to automatically find the most appropriate distribution of emitter types depending on the situation. \algoname{} is summarised in Fig.\ref{fig:concept} and Algo.\ref{algo:ME-MAP-Elites}.

\subsection{Emitter definitions}
The different emitter types are designed to be specialised on a different aspect of the optimisation process. Some of them favour the exploration of the behavioural space (maximising diversity), while others focus on improving the performance of the solutions contained in the archive (maximising quality). Finally, some of them follow different quality/diversity trade-offs. 
We use four different types of emitters. Three of them are directly taken from the CMA-ME algorithm \citep{fontaine2020covariance}, while the last-one captures the selection mechanism of MAP-Elites into an emitter. 

\begin{itemize}[leftmargin=*]
    \item \textbf{The optimising emitter} rewards solutions with a high fitness value (similar to the original CMA-ES algorithm). It returns at each generation the current population of the CMA-ES process which will be used as a batch of solutions to be evaluated. The CMA-ES process is initialised by randomly selecting a solution in the archive, which serves as the initial mean of the sampling distribution (with a fixed variance). This emitter is taken from CMA-ME. 
    \item \textbf{The random direction emitter} rewards solutions that move along a predefined direction in the behavioural space. The direction is defined according to the behavioural descriptor of a randomly selected solution of the archive and a randomly generated vector representing an orientation in the behavioural descriptor space. This emitter is also based on CMA-ES and is taken from CMA-ME.
    \item \textbf{The improvement emitter} rewards solutions that improve the archive. The archive improvement is defined as the fitness of the solution (assumed to be strictly positive) when a new cell is filled, or by the fitness improvement when a generated solution replaces an existing one. As a consequence, this emitter moves in the direction that generates either new or better solutions. This emitter is also based on CMA-ES and is taken from CMA-ME.
    \item \textbf{The random emitter} is based on vanilla MAP-Elites and generates a batch of solutions by applying a variation operator to a set of solutions randomly selected from the archive. We use the ``directional variation'' operator \cite{vassiliades2018discovering}, which has shown to provide better results in our preliminary experiments (not shown), than a polynomial mutation operator and a simulated binary crossover (SBX, \cite{deb2006multi}) operator.
\end{itemize}

\subsection{Main steps}
At every generation, a predefined number of emitters (e.g.,  12) can be active. These active emitters are selected from a heterogeneous pool of emitters in which each emitter type is represented at least as many times as there are active emitter slots. For instance, if four emitter types are considered for 12 active slots, then the pool will contain 12 instances of each emitter type, for a total of 48 emitters. This construction of the emitter pool enables the algorithm to select exclusively one type of emitter if this type is assessed to be the most effective one. This makes the algorithm capable of automatically becoming equivalent to one of the CMA-ME variants (i.e, using exclusively one of the CMA-ES based emitter type), or to MAP-Elites (i.e., using exclusively the random emitter). 

When an emitter becomes active, it remains active as long as no stopping criterion is reached. This allows emitters to execute multiple optimisation steps before being potentially replaced by another emitter. This is important for emitters using internal optimisation processes with internal states, such as those based on CMA-ES (which depend on the state of the Gaussian distribution used to sample the population).  When a stopping criterion is reached, the emitter is removed from the set of active emitters and returns to the emitter pool with its internal state reset. For the CMA-ES based emitters, the stopping criterion are those defined in the CMA-ES algorithm (configured with the default parameter values provided in Nikolaus Hansen's 2013 C implementation) or if none of the sampled solutions is added to the archive during one generation. The original implementation of CMA-ME~\cite{fontaine2020covariance} uses a different stopping criterion for the optimising emitter, which is run until convergence. In this paper, we decided to adopt a common definition of the stopping criterion for all our CMA-ES based emitters. 
The random emitter is systematically deactivated and returned to the pool after each generation, as its execution does not depend on an internal state. Because all the active emitters do not finish at the same time, only a subset of the active emitters needs to be replaced at every generation (i.e., those that have reached a stopping criterion).

\subsection{Emitter selection with bandit algorithm}
The selection of the active emitters out of the emitter pool uses Upper Confidence Bound - 1 algorithm (UCB1), a bandit selection algorithm that minimises the expected regret by balancing high-predicted reward and uncertainty \citep{auer2002finite, garivier2011upper}. A similar selection approach has been used in \cite{gaier2020automating} to select among different mutation operators on a learned encoding of solutions. The goal of the bandit algorithm is to select the option (here the emitter) with the highest expected reward. Here, the reward corresponds to the proportion of solutions added to the archive out of solutions sampled by this emitter. 
UCB1 selects the option with the highest potential reward: 
\begin{equation}
I = \argmax_e R(e)+\zeta\sqrt{\log(t)/(N_t(e))}
\label{eq:ucb}
\end{equation}
where $R(e)$ is the empirical mean reward of emitter $e$, $t$ is the total number of selection rounds done so far, $\zeta$ is a hyper-parameter, and $N_t(e)$ represents the number of times option $e$ has been selected so far. The second term of the equation captures the uncertainty of the predictions, based on how frequently an emitter is selected. This formulation balances exploitation and exploration as a weighted sum of the empirical mean reward and the uncertainty of an emitter. 

In addition to the reward being stochastic, like in the usual  multi-armed bandit scenario, the context considered in this paper is also non-stationary (the best emitters might change over time) and allows for multiple plays per turn (multiple emitters are selected every generation). These conditions are largely discussed in the literature of bandit algorithms \cite{ uchiya2010algorithms, besbes2014stochastic,garivier2011upper}.
Consequently, we adapted UCB1 by basing our selection only on the data collected over the last 50 generations, as suggested in \cite{garivier2011upper, gaier2020automating}, and by selecting not only the best emitter but the $nb\_needed\_emitters$ best emitters following equation~\ref{eq:ucb} (being the number of emitters to be selected at each generation). This formulation enables the algorithm to select the most appropriate set of emitters at a given time and to adapt its decisions periodically (i.e., when an emitter is reset) by taking into account only recent data.  

Other multi-armed bandit algorithms can be used instead of UCB1. For instance, we successfully tested the non-stationary and multiple-plays variants of the Exp3 algorithm (Exp3.M, \cite{uchiya2010algorithms}, and Rexp3 \cite{besbes2014stochastic}). However, given that UCB1 was significantly easier to implement while offering the same level of performance, we decided to use UCB1 for all our experiments.

\section{Experimental evaluation}
The performance of \algoname{} is evaluated on six simulated tasks: 1) ``Rastrigin-proj'', 2) ``Rastrigin-multi'', 3) ``Sphere'', 4) ``Redun-dant-arm'', 5) ``Hexapod-uni'', 6) ``Hexapod-omni'', described in the following sections.
%\subsection{Experimental setups}
\subsubsection*{Rastrigin-proj and Rastrigin-multi}
The first two tasks consider the Rastrigin function with two different definitions for the behavioural descriptor \citep{muhlenbein1991parallel}. Closely following the definitions introduced in CMA-ME \cite{fontaine2020covariance}, the search space is defined using 100 dimensions and the extremum of Rastrigin is shifted to $x^*_i = 0.4*5.12 ,   \forall i$.
\begin{equation*}
    \textrm{fit}(\mathbf{x}) = \sum_{i=1}^{n}{[ (x_i-0.4*5.12)^2 - 10 \cos(2\pi (x_i-0.4*5.12))]}
\end{equation*}

The first task, called thereafter ``Rastrigin-proj'', uses the behavioural descriptor defined in \cite{fontaine2020covariance} in which every dimension of $\mathbf{x}$ is projected to contribute to the first or second dimension of the behavioural descriptor:
\begin{equation*}
\begin{aligned}
\textrm{bd}_{\textrm{proj}}(\mathbf{x}) =\left( \sum_{i=1}^{\floor{\frac{n}{2}}}{\textrm{clip}(x_i) } , \sum_{i=\floor{\frac{n}{2}}+1}^{n}{\textrm{clip}(x_i) }  \right) \\
\textrm{clip}(x_i) = \begin{cases}
               x_i, & \textrm{if } -5.12\leq x_i \leq 5.12\\
               5.12/x_i,& \textrm{otherwise}
            \end{cases}
\end{aligned}
\end{equation*}
The objective of this behavioural descriptor definition is to create multiple domains in the search space capable of filling the entire collection, while not all the domains being equivalent in terms of fitness. The challenge for the QD algorithm thus becomes to find the right domain to create the largest collection with the best fitness. More precisely, solutions outside the [$-5.12$, $5.12$] range are penalised. The library used in our experiments (detailed below) requires the search space to be bounded. Therefore, to maintain the challenging aspect of this task, the bounds of the space have been set to [$-51.2$, $51.2$].

The second task, called thereafter ``Rastrigin-multi'', uses another behavioural descriptor that can be found in the literature \citep{justesen2019map} in which the behavioural descriptor is defined as  $\textrm{bd}_{\textrm{multi}}(\mathbf{x}) =\left( x_1 , x_2 \right)$. This task is called ``multi'' because the definition of the behavioural descriptor creates multiple local optima in the archive instead of a unimodal structure of the fitness unlike in the ``proj'' variant. The search space is bounded between [$-5.12$, $5.12$] like in previous works from the literature ~\cite{ali2005numerical,flageat2020fast}. 

\subsubsection*{Sphere}
The third task is also taken from \cite{fontaine2020covariance} and adopts the same definition of the genotype (100 dimensions and bounded between $-51.2$ and $51.2$) and the same behavioural descriptor as the Rastrigin-proj task ($\textrm{bd}_{\textrm{proj}}$). The fitness is built on the Sphere function with the same shift of the extremum point as introduced before:
\begin{equation*}
    \textrm{fit}(\mathbf{x}) = \sum_{i=1}^{n}{(x_i-0.4*5.12)^2}
\end{equation*}

\subsubsection*{Redundant-arm}
The last three tasks are taken from \cite{cully2018quality}. The Redundant-arm task considers a planar robotic arm with 100 degrees of freedom (only height in \cite{cully2018quality}), which can move between $-\pi$ and $\pi$ radians. The behavioural descriptor is the Cartesian position of the robot's gripper: $\textrm{bd}(\mathbf{x}) =\left(x_{\textrm{gripper}},y_{\textrm{gripper}} \right)$, while the fitness is the opposite of variance of the joint's articular position: $\textrm{fit}(\mathbf{x}) = - \mathrm{Var}(\mathbf{x}) $
This fitness definition encourages every degree of freedom to contribute equally to the movement of the arm.

\subsubsection*{Hexapod-uni and Hexapod-omni}
These two tasks are based on a simulated hexapod introduced in \cite{cully2015robots} and reused in multiple works in the literature \citep{cully2018quality,vassiliades2018discovering,gaier2020automating}. The hexapod has 12 directly controlled degrees of freedom, which are independently governed by a sin-wave like controller parametrised by its amplitude, phase and duty cycle. This leads to a total of 36 parameters with values bounded between 0 and 1 which are then scaled according to the admissible range of each joint. More details can be found in \cite{cully2015robots}.
The differences between the Hexapod-uni and Hexapod-omni tasks are the fitness and behavioural descriptor definitions. The first one considers a unidirectional locomotion tasks in which the robot has to learn to walk as fast as possible on a straight line, exactly following the definitions from \cite{cully2015robots}. This is encoded in the fitness function as the $x_{\textrm{pos}}$ Cartesian coordinate of the robot after walking during 5 seconds. The behavioural descriptor is a six-dimensional vector corresponding to the proportion of time that each leg spends in contact with the ground. More details in \cite{cully2015robots}.

Finally, Hexapod-omni is an omnidirectional locomotion task in which the robot has to learn to walk in every direction. The behavioural descriptor is the $(x_{\textrm{pos}},y_{\textrm{pos}})$  Cartesian coordinates of the robot after walking during 3 seconds. The fitness, taken from \cite{cully2015evolving} is designed to encourage the robot to follow circular trajectories. It is defined as the absolute difference between the robot’s final orientation and the tangent of the ideal circular trajectory. More details can be found in \cite{cully2015evolving}.

\subsection{Compared algorithms}
We compare six algorithms (or variants) on the six tasks described above: 1) \emph{CMA-ME opt}, 2) \emph{CMA-ME dir}, 3) \emph{CMA-ME imp}, 4) \emph{MAP-Elites}, 5) \emph{\algoname{} uniform}, 6) \emph{\algoname{} UCB}. The CMA-ME variants are directly taken from \cite{fontaine2020covariance} and use a single type of emitter, respectively the optimiser emitter, the random direction emitter, and the improvement emitter. 
In the \emph{\algoname{} uniform} variant, the emitter pool is fixed and composed of three emitters for each of the four emitter types. The \emph{\algoname{} UCB} variant dynamically adapts set of active emitters using the UCB1 algorithm as described above.  

\subsection{Implementation and Hyper-parameters}
For a fair and consistent comparison, all the compared algorithms use the same hyper-parameter values. In particular, each experiment is replicated 20 times, during 20k generations, with 12 active emitter slots for a total batch size of 600 (50 per active emitters). UCB1 uses $\zeta = 0.05$, and MAP-Elites uses the following values for the line operator:$\sigma_{\textrm{line}}=0.1$ and $\sigma_{\textrm{iso}}=0.01$. The CMA-ES based emitters use an initial $\sigma=0.5$ in the Rastrigin and Sphere tasks, like in the original paper~\cite{fontaine2020covariance}), and $\sigma=0.25$ in the other tasks because the bounded search space is smaller in these last experiments. The number of randomly generated solutions used for initialisation of the grids is set to $100$.

It is important to note that in our experiments, the search space is bounded, and any generated solution will be clipped to fit in the search space. More advanced and better approaches exist to handle solutions outside the admissible range, in particular when applied with CMA-ES~\cite{biedrzycki2020handling}. However, we leave the investigation of alternative clipping methods for future works. 

All the experiments use a grid with a resolution of $100$x$100$, except the Hexapod-uni task, which use the same grid as in Cully et al. \cite{cully2015robots} with $5^6$ cells. We use these resolutions because MAP-Elites usually performs well in this configuration, while struggling when the collection size exceeds 20k solutions. Fontaine et al.~\cite{fontaine2020covariance} evaluated CMA-ME in larger grid ($512$x$512$) and demonstrated its capability to scale to large collection size. In this paper, we want to compare \algoname{} and other algorithms in domains where MAP-Elites is well established~\cite{cully2018quality}.

Our implementation is based on the Sferes$_{v2}$ library \citep{Mouret2010} and the Quality-Diversity framework introduced by Cully and Demiris~\cite{cully2018quality}. The hexapod experiments use the Dart simulator \citep{Lee2018}. It is important to note that we re-implemented the CMA-ME algorithms in C++ (the original one was in C\#). While we did our best to follow the original implementation as closely as possible, some discrepancies might still exist. Yet, the relative performance of all the variants and MAP-Elites have been reproduced in the experiments below. One can note that the original implementation of CMA-ME also allows the use of heterogeneous sets of emitters, even though it was not studied in the paper~\cite{fontaine2020covariance}. The source code of our implementation and a singularity container \citep{kurtzer2017singularity}, containing the compilation and runtime environment for instantaneous replication, can be found at \url{https://github.com/adaptive-intelligent-robotics/Multi-Emitter_Map-Elites}

\subsection{Metrics}

All the algorithms are evaluated following the same procedure and metrics. In particular, we report the evolution of the archive size, the highest fitness present in the archive, and QD-Score over the number of generations. The QD-Score, introduced in \cite{pugh2015confronting}, is the sum of the fitness of all the solutions contained in the archive. For easier comparison, the fitness values are normalised between $0$ and $1$, respectively the worst and best possible fitness value given the bounded search space (considering [$-5.12$, $5.12$] and otherwise clipped to 0 for the two Rastrigin tasks and Sphere like in CMA-ME~\cite{fontaine2020covariance}). The only exception is the Hexapod-uni as its maximal walking distance is unknown.

\subsection{Results}

\begin{figure*}[!t]
\includegraphics[width=0.97\textwidth]{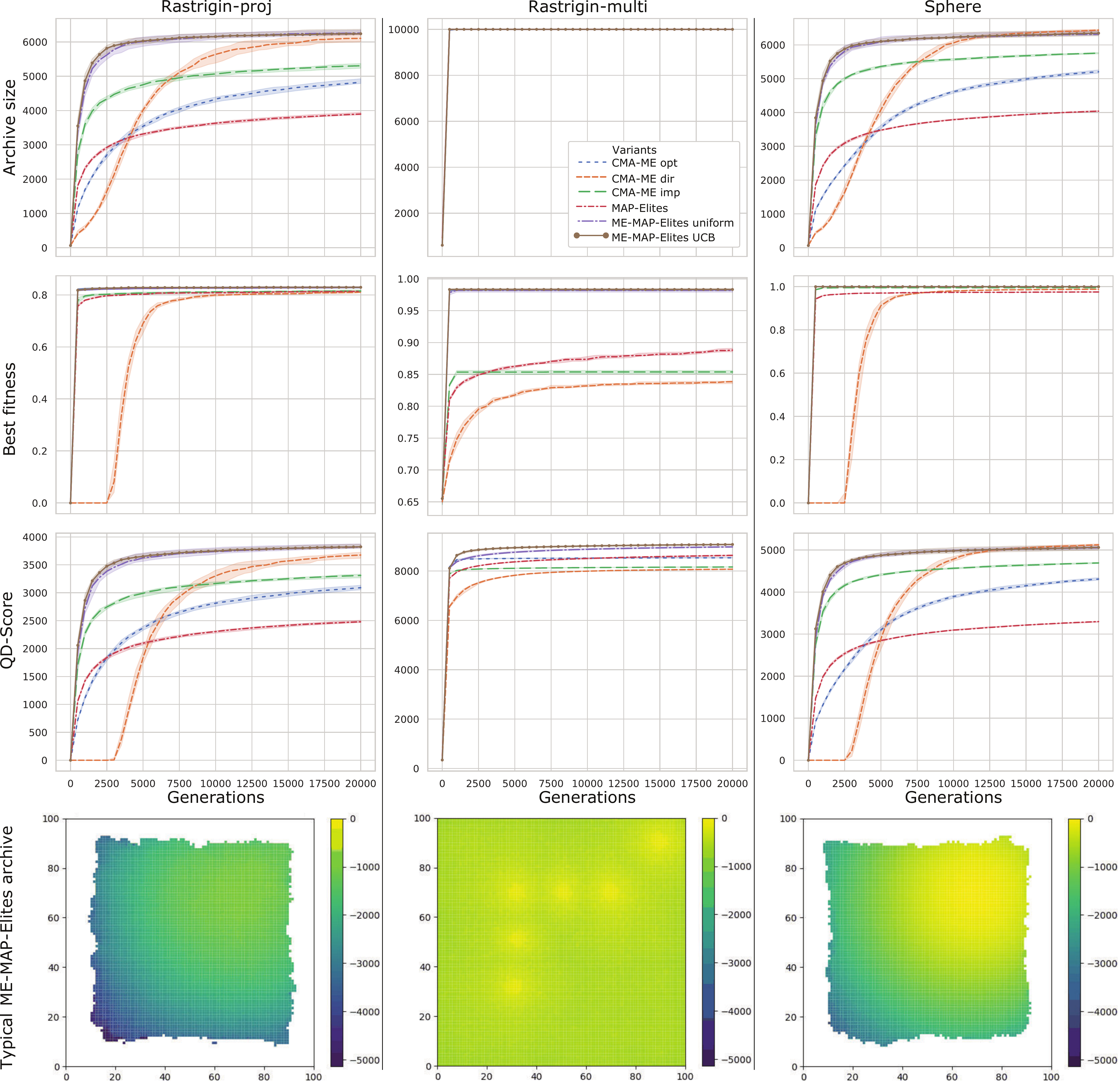}
\caption{Results for the Rastrigin-proj, Rastrigin-multi, and Sphere experiments. For each experiment, the progression of the archive size, fitness of the best individual in the archive, QD-Score, and a typical archive obtained with ME-MAP-Elites is displayed. Each experiment has been replicated 20 times for 20,000 generations. The graphs represent the median as a coloured bold line, while the shaded area extends to the first and the third quartiles. (Note: the archive shown for the Rastrigin-multi task presents some variations that are not perceivable.)} 
\label{fig:result_1}
\end{figure*}

\begin{figure*}[!t]
\includegraphics[width=0.97\textwidth]{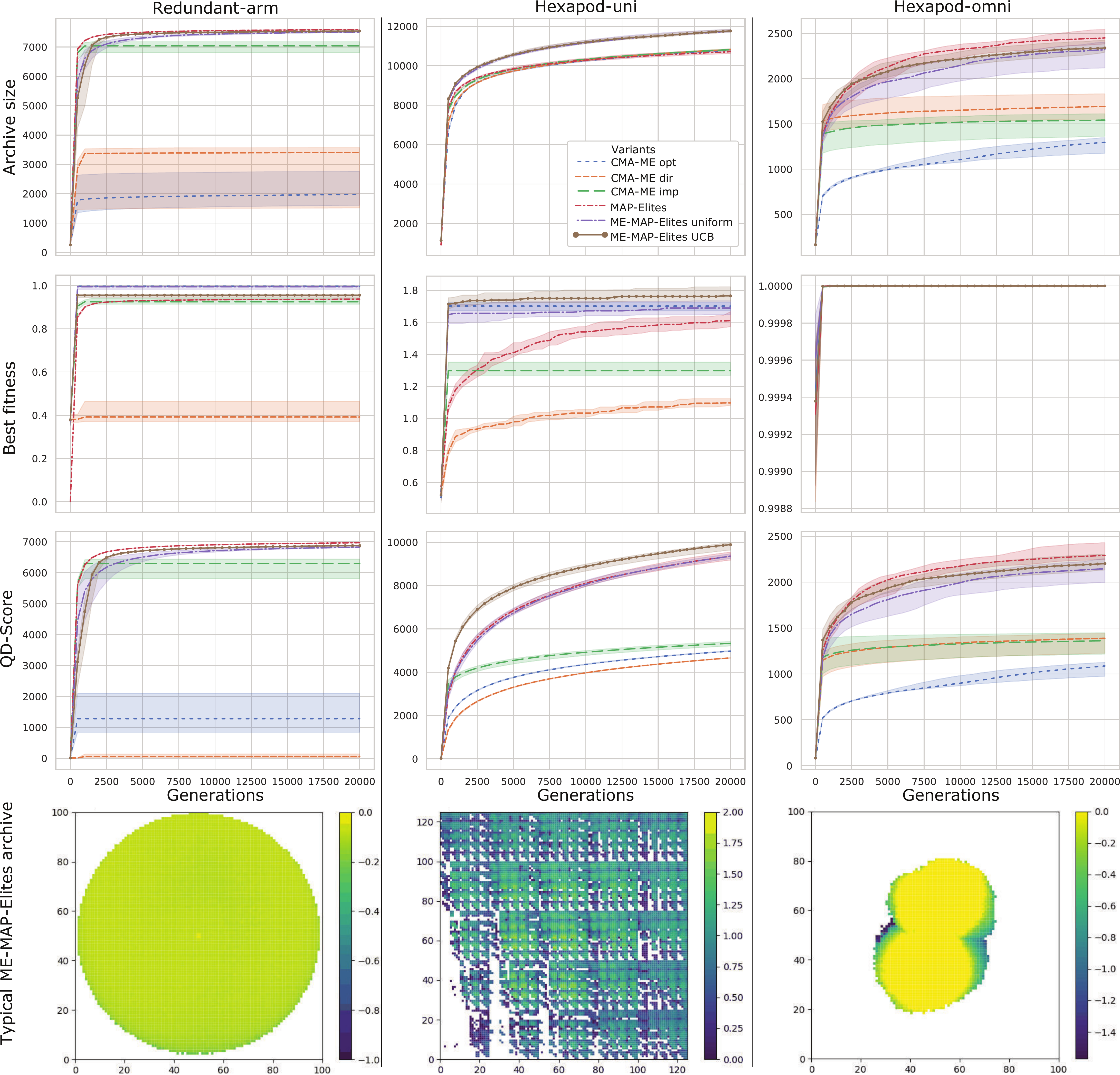}
\caption{Results for the Redundant-arm, Hexapod-uni, and Hexapod-omni. This figure uses the same notations as Fig.~\ref{fig:result_1}.}
\label{fig:result_2}
\end{figure*}

Fig.\ref{fig:result_1} and \ref{fig:result_2} show the results of each algorithm on the different tasks. The first thing that can be noted is that each emitter type displays, as highlighted in the CMA-ME paper~\cite{fontaine2020covariance}, a different trade-off between improvement of the quality (via local optimisation) or improvement of the diversity (via the exploration of the behavioural space). The \emph{optimising emitter} is designed to generate solutions with the highest possible fitness and we can observe in the experimental results that the \emph{CMA-ME opt} variant is systematically higher in terms of best fitness than the other CMA-ME variants. Similarly, the \emph{random direction emitter} aims at exploring the behavioural space by design, and this characteristic is demonstrated by the fact that \emph{CMA-ME dir} has the largest archive size out of the three CMA-ME variants in all the tested cases, except in the Redundant-arm experiment.
Finally, the \emph{improvement emitter} is defined to improve either the archive size or the fitness of the solutions and we can observe that  \emph{CMA-ME imp} is systematically ranked between the two other CMA-ME variants, except for the Redundant-arm and the Hexapod-uni experiments in which \emph{CMA-ME imp} obtains better results on the archive size and the QD-score.

The selection and mutation operators of \emph{MAP-Elites} make its progress relatively unbiased to a specific aspect (archive size, fitness or QD-score). It can be observed that it outperforms all the CMA-ME variants in terms of QD-Score in the tasks with limited search space, but shows limited capabilities in the Rastrigin-proj and the Sphere tasks. Moreover, the data efficiency of \emph{MAP-Elites} is several orders of magnitude slower than the CMA-ME variants in some cases. For instance, in the Rastrigin-multi experiment, it takes more than 10k generations for MAP-Elites to reach the same QD-Score as the one obtained by \emph{CMA-ME imp} after 500 generations. Similarly, in the Sphere experiment, the QD-score of \emph{MAP-Elites} after 20k generations is obtained in less than 500 generations with \emph{CMA-ME imp}. Conversely, we can note that \emph{MAP-Elites} is the top performing algorithm, in terms of archive size and QD-score, for both the Redundant-arm and Hexapod-omni experiments. This result is most likely linked with the directional variation operator being particularly effective in these tasks. Indeed, most weighted averages of two optimal solutions in the Redundant-arm experiment will lead to a different optimal solution. Overall, these results confirm the conclusions of the CMA-ME paper~\cite{fontaine2020covariance} regarding the benefits of emitters compared to MAP-Elites and extend it to additional domains.

The two \algoname{} variants introduced in this paper take the best of both worlds by combining the strengths of all emitter types. The final QD-Scores of \emph{\algoname{} UCB} are systematically either better by a certain margin or equivalent to all the CMA-ME variants and \emph{MAP-Elites}. The higher QD-Score of \emph{\algoname{} UCB} compared to the CMA-ME variants and \emph{MAP-Elites} are all statistically significant (all p-values $< 1.5e^{-5}$, computed with Wilcoxon rank-sum test). The only exceptions are for the Hexapod-omni and the Sphere experiments in which \emph{\algoname{} UCB} is outperformed by \emph{MAP-Elites} and  \emph{CMA-ME imp} respectively. However, these differences are inconclusive after the Holm-Bonferroni correction \citep{shaffer1995multiple} (p-values equal $0.08$ and $0.001$ respectively). The Redundant-arm is the only experiment in which \emph{MAP-Elites} outperforms \emph{\algoname{} UCB} with a statistically significant difference (p-values $< 1.5e^{-5}$).

The \emph{\algoname{} uniform} variant is a simpler implementation of the proposed algorithm, as the set of active emitters remains the same and no bandit algorithm is used. It is interesting to note that this variant is often ranked second or third (in terms of QD-score), just after the \emph{UCB} variant and the closest competitor. While the final QD-score values for the \emph{uniform} variant remains very close to the \emph{UCB} variant and not always statistically significant, the main strength of the \emph{UCB} variant is to reach these final values faster. This is particularly noticeable in the Redundant-arm experiment, where the final QD-score of the \emph{uniform} variant is reached in only half of the number of generations by the \emph{UCB} variant.

In the worst cases, on all the metrics and all the considered tasks, \emph{\algoname{} UCB} is equivalently good as the best competitor. 
Moreover, the proposed algorithm is the only one to show this consistency. While \emph{MAP-Elites} performs excellently on two tasks (the Redundant-arm and the Hexapod-oni), it dramatically fails on two others (the Sphere and Rastrigin-proj). The same behaviour can be observed with the CMA-ME variants, which demonstrate superb performance in some tasks, but never in all of them.

\begin{figure}[!b]
\includegraphics[width=0.96\columnwidth]{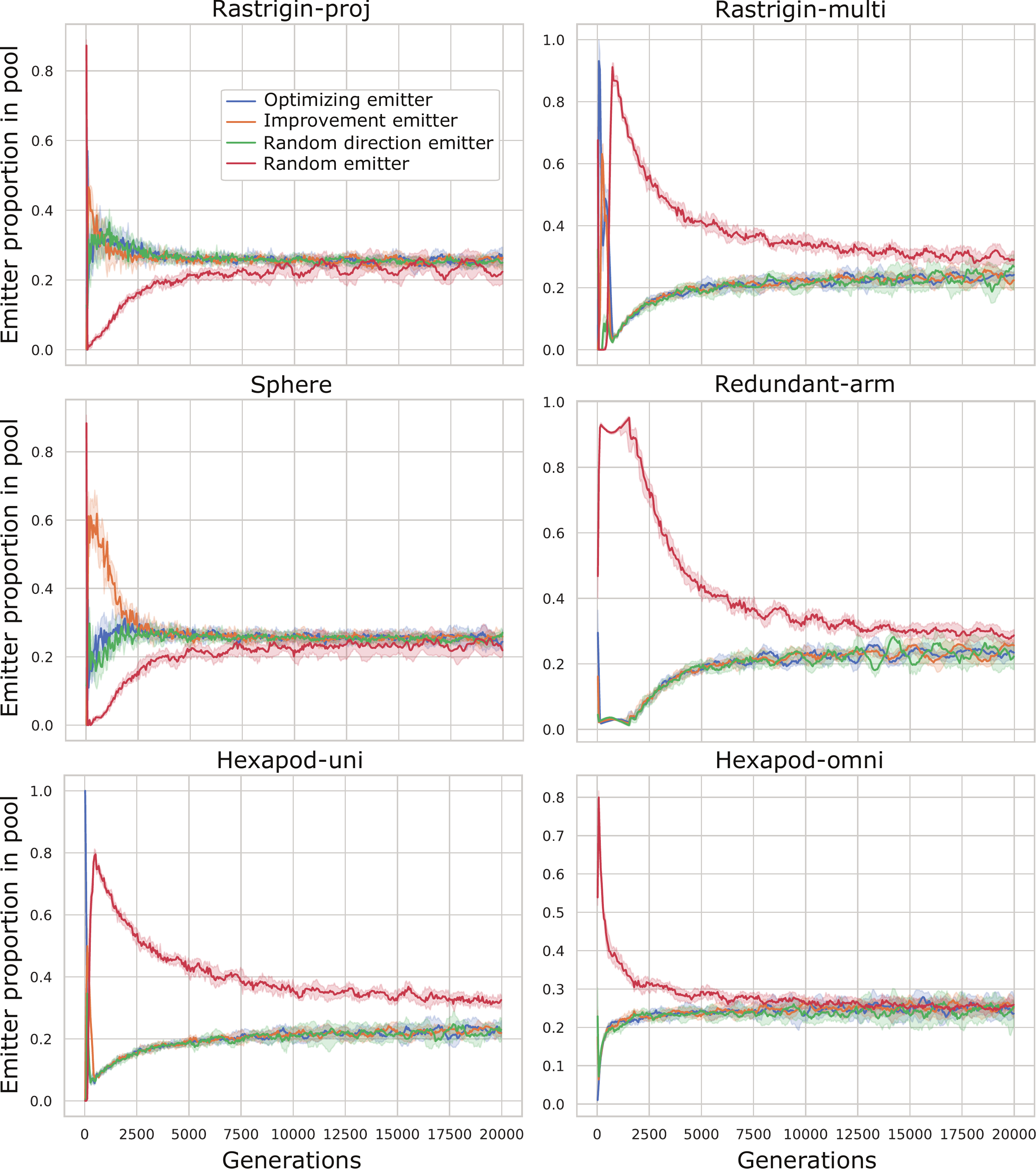}
\caption{Proportion of each emitter type in the set of active emitters over the number of generations. To improve readability, the data has been smoothed with a triangular sliding averaging window of width 50 generations. Like in the previous figures, each experiment has been replicated 20 times and the resulting values are displayed as the median with a bold coloured line and with shared areas that extend to the first and third quartiles.}
\label{fig:mab}
\end{figure}

Fig.~\ref{fig:mab} shows the distribution of the emitter types in the active emitter set over the number of generations. We can see that while there is a lot of variation during the first 2500 generations, the selection of emitters usually converges toward a uniform selection. This confirms the results observed above indicating that at the end of the experiments the \emph{uniform} and \emph{UCB} variants perform similarly. Therefore, the key difference in performance during the optimisation process originates from the first 2500 generations. We can observe that depending on the experiment, a different type of emitter is predominant during this period. For instance, in the Sphere experiment, the \emph{random direction emitter} is predominant, while in the Redundant-arm, the predominant emitter is the \emph{random emitter}. It is interesting to note that these two emitter types correspond respectively to the best variant in each of these experiment. This indicates that the bandit algorithm successfully selects the most appropriate emitter type, while maintaining the capability of dynamically changing its selection over the duration of the experiment, leading to better overall performance.

\section{Conclusion and discussion}
In this paper, we introduced \algoname{} UCB, a direct extension of CMA-ME, which combines the use of a heterogeneous set of emitters with a bandit algorithm (UCB1) to improve the quality and diversity of the produced archives with a better data efficiency. The experimental evaluation presented in this paper shows that this algorithm is systematically either significantly better or equivalent to the best of the compared algorithms, which is task-dependent. 
In our implementation, we exclusively used emitter types that were introduced in \cite{fontaine2020covariance} and a new emitter to reproduce MAP-Elites's behaviour. However, it would be interesting to investigate in future work additional types of emitters, for instance, based on a Data-driven representation of the solution, like in \cite{gaier2020automating} and \cite{gaier2018data}. It would also be interesting to see how the concept of multi-emitter can be combined with the SAIL algorithm as they are two orthogonal research directions to improve the data-efficiency of QD algorithms.

\begin{acks}
This work was supported by the Engineering and Physical Sciences Research Council (EPSRC) grant EP/V006673/1 project REcoVER. I would like to thank the members of the Adaptive and Intelligent Robotics Lab for their very valuable comments.  
\end{acks}

%

%% file: main.bbl
\begin{thebibliography}{}

\bibitem[Ali et~al., 2005]{ali2005numerical}
Ali, M.~M., Khompatraporn, C., and Zabinsky, Z.~B. (2005).
\newblock A numerical evaluation of several stochastic algorithms on selected
  continuous global optimization test problems.
\newblock {\em Journal of global optimization}, 31(4):635--672.

\bibitem[Alvarez et~al., 2019]{alvarez2019empowering}
Alvarez, A., Dahlskog, S., Font, J., and Togelius, J. (2019).
\newblock Empowering quality diversity in dungeon design with interactive
  constrained map-elites.
\newblock {\em IEEE Conference on Games (CoG)}.

\bibitem[Auer et~al., 2002]{auer2002finite}
Auer, P., Cesa-Bianchi, N., and Fischer, P. (2002).
\newblock Finite-time analysis of the multiarmed bandit problem.
\newblock {\em Machine learning}, 47(2-3):235--256.

\bibitem[Besbes et~al., 2014]{besbes2014stochastic}
Besbes, O., Gur, Y., and Zeevi, A. (2014).
\newblock Stochastic multi-armed-bandit problem with non-stationary rewards.
\newblock In {\em Advances in neural information processing systems}, pages
  199--207.

\bibitem[Biedrzycki, 2020]{biedrzycki2020handling}
Biedrzycki, R. (2020).
\newblock Handling bound constraints in cma-es: An experimental study.
\newblock {\em Swarm and Evolutionary Computation}, 52:100627.

\bibitem[Bongard et~al., 2006]{bongard2006resilient}
Bongard, J., Zykov, V., and Lipson, H. (2006).
\newblock Resilient machines through continuous self-modeling.
\newblock {\em Science}, 314(5802):1118--1121.

\bibitem[Cully, 2019]{cully2019Autonomous}
Cully, A. (2019).
\newblock Autonomous skill discovery with quality-diversity and unsupervised
  descriptors.
\newblock In {\em Proceedings of the Genetic and Evolutionary Computation
  Conference}, pages 81--89. ACM.

\bibitem[Cully et~al., 2015]{cully2015robots}
Cully, A., Clune, J., Tarapore, D., and Mouret, J.-B. (2015).
\newblock Robots that can adapt like animals.
\newblock {\em Nature}, 521(7553):503--507.

\bibitem[Cully and Demiris, 2018]{cully2018quality}
Cully, A. and Demiris, Y. (2018).
\newblock Quality and diversity optimization: A unifying modular framework.
\newblock {\em IEEE Transactions on Evolutionary Computation}, 22(2):245--259.

\bibitem[Cully and Mouret, 2015]{cully2015evolving}
Cully, A. and Mouret, J.-B. (2015).
\newblock Evolving a behavioral repertoire for a walking robot.
\newblock {\em Evolutionary Computation}.

\bibitem[Deb et~al., 2006]{deb2006multi}
Deb, K., Sinha, A., and Kukkonen, S. (2006).
\newblock Multi-objective test problems, linkages, and evolutionary
  methodologies.
\newblock In {\em Proceedings of the 8th annual conference on Genetic and
  evolutionary computation}, pages 1141--1148.

\bibitem[Evans and Gao, 2016]{evans2016deepmind}
Evans, R. and Gao, J. (2016).
\newblock Deepmind ai reduces google data centre cooling bill by 40\%.
\newblock {\em DeepMind blog}, 20:158.

\bibitem[Flageat and Cully, 2020]{flageat2020fast}
Flageat, M. and Cully, A. (2020).
\newblock Fast and stable map-elites in noisy domains using deep grids.
\newblock In {\em Artificial Life Conference Proceedings}, pages 273--282. MIT
  Press.

\bibitem[Fontaine et~al., 2019]{fontaine2019}
Fontaine, M.~C., Lee, S., Soros, L.~B., Silva, F. D.~M., Togelius, J., and
  Hoover, A.~K. (2019).
\newblock Mapping hearthstone deck spaces with map-elites with sliding
  boundaries.
\newblock In {\em Proceedings of The Genetic and Evolutionary Computation
  Conference}. ACM.

\bibitem[Fontaine et~al., 2020]{fontaine2020covariance}
Fontaine, M.~C., Togelius, J., Nikolaidis, S., and Hoover, A.~K. (2020).
\newblock Covariance matrix adaptation for the rapid illumination of behavior
  space.
\newblock {\em Proceedings of The Genetic and Evolutionary Computation
  Conference}.

\bibitem[Gaier et~al., 2017]{gaier2017aerodynamic}
Gaier, A., Asteroth, A., and Mouret, J.-B. (2017).
\newblock Aerodynamic design exploration through surrogate-assisted
  illumination.
\newblock In {\em 18th AIAA/ISSMO Multidisciplinary Analysis and Optimization
  Conference}, page 3330.

\bibitem[Gaier et~al., 2018]{gaier2018data}
Gaier, A., Asteroth, A., and Mouret, J.-B. (2018).
\newblock Data-efficient design exploration through surrogate-assisted
  illumination.
\newblock {\em Evolutionary computation}, pages 1--30.

\bibitem[Gaier et~al., 2019]{gaier2019quality}
Gaier, A., Asteroth, A., and Mouret, J.-B. (2019).
\newblock Are quality diversity algorithms better at generating stepping stones
  than objective-based search?
\newblock In {\em Proceedings of the Genetic and Evolutionary Computation
  Conference Companion}, pages 115--116.

\bibitem[Gaier et~al., 2020]{gaier2020automating}
Gaier, A., Asteroth, A., and Mouret, J.-B. (2020).
\newblock Discovering representations for black-box optimization.
\newblock {\em Proceedings of the 2020 Genetic and Evolutionary Computation
  Conference.}

\bibitem[Garivier and Moulines, 2011]{garivier2011upper}
Garivier, A. and Moulines, E. (2011).
\newblock On upper-confidence bound policies for switching bandit problems.
\newblock In {\em International Conference on Algorithmic Learning Theory},
  pages 174--188. Springer.

\bibitem[Hansen, 2016]{hansen2016cma}
Hansen, N. (2016).
\newblock The cma evolution strategy: A tutorial.
\newblock {\em arXiv preprint arXiv:1604.00772}.

\bibitem[Hansen and Ostermeier, 2001]{hansen2001completely}
Hansen, N. and Ostermeier, A. (2001).
\newblock Completely derandomized self-adaptation in evolution strategies.
\newblock {\em Evolutionary computation}, 9(2):159--195.

\bibitem[Justesen et~al., 2019]{justesen2019map}
Justesen, N., Risi, S., and Mouret, J.-B. (2019).
\newblock Map-elites for noisy domains by adaptive sampling.
\newblock In {\em Proceedings of the Genetic and Evolutionary Computation
  Conference Companion}, pages 121--122. ACM.

\bibitem[Komorowski et~al., 2018]{komorowski2018artificial}
Komorowski, M., Celi, L.~A., Badawi, O., Gordon, A.~C., and Faisal, A.~A.
  (2018).
\newblock The artificial intelligence clinician learns optimal treatment
  strategies for sepsis in intensive care.
\newblock {\em Nature medicine}, 24(11):1716--1720.

\bibitem[Kurtzer et~al., 2017]{kurtzer2017singularity}
Kurtzer, G.~M., Sochat, V., and Bauer, M.~W. (2017).
\newblock Singularity: Scientific containers for mobility of compute.
\newblock {\em PloS one}, 12(5).

\bibitem[Lee et~al., 2018]{Lee2018}
Lee, J., Grey, M.~X., Ha, S., Kunz, T., Jain, S., Ye, Y., Srinivasa, S.~S.,
  Stilman, M., and Liu, C.~K. (2018).
\newblock {DART}: Dynamic animation and robotics toolkit.
\newblock {\em The Journal of Open Source Software}, 3(22):500.

\bibitem[Lehman and Stanley, 2011a]{lehman2011abandoning}
Lehman, J. and Stanley, K.~O. (2011a).
\newblock Abandoning objectives: Evolution through the search for novelty
  alone.
\newblock {\em Evolutionary computation}, 19(2):189--223.

\bibitem[Lehman and Stanley, 2011b]{lehman2011evolving}
Lehman, J. and Stanley, K.~O. (2011b).
\newblock Evolving a diversity of virtual creatures through novelty search and
  local competition.
\newblock In {\em Proceedings of the 13th annual conference on Genetic and
  evolutionary computation}, pages 211--218. ACM.

\bibitem[Mouret and Clune, 2015]{mouret2015illuminating}
Mouret, J.-B. and Clune, J. (2015).
\newblock Illuminating search spaces by mapping elites.
\newblock {\em arXiv preprint arXiv:1504.04909}.

\bibitem[Mouret and Doncieux, 2010]{Mouret2010}
Mouret, J.-B. and Doncieux, S. (2010).
\newblock {SFERES}v2: Evolvin' in the multi-core world.
\newblock In {\em Proc. of Congress on Evolutionary Computation (CEC)}, pages
  4079--4086.

\bibitem[M{\"u}hlenbein et~al., 1991]{muhlenbein1991parallel}
M{\"u}hlenbein, H., Schomisch, M., and Born, J. (1991).
\newblock The parallel genetic algorithm as function optimizer.
\newblock {\em Parallel computing}, 17(6-7):619--632.

\bibitem[Murphy, 2014]{murphy2014disaster}
Murphy, R.~R. (2014).
\newblock {\em Disaster robotics}.
\newblock MIT press.

\bibitem[Paolo et~al., 2019]{paolo2019unsupervised}
Paolo, G., Laflaquiere, A., Coninx, A., and Doncieux, S. (2019).
\newblock Unsupervised learning and exploration of reachable outcome space.
\newblock {\em algorithms}, 24:25.

\bibitem[Pugh et~al., 2015]{pugh2015confronting}
Pugh, J.~K., Soros, L., Szerlip, P.~A., and Stanley, K.~O. (2015).
\newblock Confronting the challenge of quality diversity.
\newblock In {\em Proceedings of the 2015 on Genetic and Evolutionary
  Computation Conference}, pages 967--974. ACM.

\bibitem[Rasmussen, 2003]{rasmussen2003gaussian}
Rasmussen, C.~E. (2003).
\newblock Gaussian processes in machine learning.
\newblock In {\em Summer School on Machine Learning}, pages 63--71. Springer.

\bibitem[Shaffer, 1995]{shaffer1995multiple}
Shaffer, J.~P. (1995).
\newblock Multiple hypothesis testing.
\newblock {\em Annual review of psychology}, 46(1):561--584.

\bibitem[Siciliano and Khatib, 2016]{siciliano2016springer}
Siciliano, B. and Khatib, O. (2016).
\newblock {\em Springer handbook of robotics (2nd edition)}.
\newblock Springer.

\bibitem[Uchiya et~al., 2010]{uchiya2010algorithms}
Uchiya, T., Nakamura, A., and Kudo, M. (2010).
\newblock Algorithms for adversarial bandit problems with multiple plays.
\newblock In {\em International Conference on Algorithmic Learning Theory},
  pages 375--389. Springer.

\bibitem[Urquhart et~al., 2019]{urquhart2019quantifying}
Urquhart, N., Hart, E., and Hutcheson, W. (2019).
\newblock Quantifying the effects of increasing user choice in map-elites
  applied to a workforce scheduling and routing problem.
\newblock In {\em International Conference on the Applications of Evolutionary
  Computation (Part of EvoStar)}, pages 49--63. Springer.

\bibitem[Vassiliades et~al., 2018]{vassiliades2018using}
Vassiliades, V., Chatzilygeroudis, K., and Mouret, J.-B. (2018).
\newblock Using centroidal voronoi tessellations to scale up the
  multidimensional archive of phenotypic elites algorithm.
\newblock {\em IEEE Transactions on Evolutionary Computation}, 22(4):623--630.

\bibitem[Vassiliades and Mouret, 2018]{vassiliades2018discovering}
Vassiliades, V. and Mouret, J.-B. (2018).
\newblock Discovering the elite hypervolume by leveraging interspecies
  correlation.
\newblock {\em Proceedings of the Genetic and Evolutionary Computation
  Conference}.

\end{thebibliography}
